\documentclass{article} 
\usepackage[preprint]{colm2026_conference}

\usepackage{microtype}
\usepackage{hyperref}
\usepackage{url}
\usepackage{booktabs}

\usepackage[utf8]{inputenc}
\usepackage[T1]{fontenc}
\usepackage{hyperref}
\usepackage{url}
\usepackage{booktabs}
\usepackage{amsfonts}
\usepackage{nicefrac}
\usepackage{microtype}
\usepackage{xcolor}

\usepackage{multicol}
\usepackage{latexsym}
\usepackage{amsmath,amsfonts,bm}

\usepackage{wrapfig,lipsum,booktabs}
\usepackage{graphicx}

\usepackage{enumitem}

\usepackage{multirow}
\usepackage{subfigure}
\usepackage{subcaption}
\usepackage{float}

\usepackage{pifont}

\usepackage{listings}
\usepackage[table]{xcolor}

\usepackage{booktabs}
\usepackage{tabularx}
\usepackage{array}
\usepackage{adjustbox}

\usepackage{amssymb}

\newcommand{\UIUC}{\textsuperscript{\ensuremath{\spadesuit}}}
\newcommand{\UWM}{\textsuperscript{\ensuremath{\heartsuit}}}
\newcommand{\Baylor}{\textsuperscript{\ensuremath{\clubsuit}}}
\newcommand{\equalcontrib}{\textsuperscript{*}}

\usepackage{fontawesome5}
\usepackage{lineno}

\definecolor{darkblue}{rgb}{0, 0, 0.5}
\hypersetup{colorlinks=true, citecolor=darkblue, linkcolor=darkblue, urlcolor=darkblue}

\title{AgentSPEX: An Agent SPecification and EXecution Language}

\author{
\shortstack[l]{%
\bfseries
Pengcheng Wang\UIUC\equalcontrib, Jerry Huang\UIUC\equalcontrib, Jiarui Yao\UIUC\equalcontrib, Rui Pan\UIUC, Peizhi Niu\UIUC,\\
Yaowenqi Liu\UIUC, Ruida Wang\UIUC, Renhao Lu\Baylor, Yuwei Guo\UWM, Tong Zhang\UIUC
}\\[0.4em]
\UIUC University of Illinois Urbana-Champaign \hspace{0.4em}
\UWM University of Wisconsin--Madison\\[0.4em]
\Baylor Baylor College of Medicine \hspace{0.4em} \equalcontrib Equal contribution. \\[0.4em]
\texttt{\{pw29, jerry8, jiarui14\}@illinois.edu}
}

\definecolor{codebg}{RGB}{248,248,248}
\definecolor{codeframe}{RGB}{200,200,200}
\definecolor{yamlkey}{RGB}{0,100,180}
\definecolor{yamlstr}{RGB}{160,32,32}
\definecolor{yamlcmt}{RGB}{100,100,100}
\definecolor{pykey}{RGB}{0,120,0}
\definecolor{pystr}{RGB}{180,50,50}
\definecolor{pycmt}{RGB}{120,120,120}

\lstdefinelanguage{yaml}{
  keywords={name, goal, config, parameters, workflow, step, task, for_each, while, if, else, switch, call, parallel, gather, instruction, enabled_tools, save_as, in, steps, condition, then, module, max_tokens_per_step, model, variable, topic, queries, file_path},
  keywordstyle=\color{yamlkey}\bfseries,
  morestring=[b]",
  stringstyle=\color{yamlstr},
  morecomment=[l]{\#},
  commentstyle=\color{yamlcmt}\itshape,
  sensitive=true,
}

\begin{document}

\ifcolmsubmission
\linenumbers
\fi

\maketitle

\begin{abstract}
Language-model agent systems commonly rely on reactive prompting, in which a single instruction guides the model through an open-ended sequence of reasoning and tool-use steps, leaving control flow and intermediate state implicit and making agent behavior potentially difficult to control. Orchestration frameworks such as LangGraph, DSPy, and CrewAI impose greater structure through explicit workflow definitions, but tightly couple workflow logic with Python, making agents difficult to maintain and modify. In this paper, we introduce AgentSPEX, an Agent SPecification and EXecution Language for specifying LLM-agent workflows with explicit control flow and modular structure, along with a customizable agent harness. AgentSPEX supports typed steps, branching and loops, parallel execution, reusable submodules, and explicit state management, and these workflows execute within an agent harness that provides tool access, a sandboxed virtual environment, and support for checkpointing, verification, and logging. Furthermore, we provide a visual editor with synchronized graph and workflow views for authoring and inspection. We include ready-to-use agents for deep research and scientific research, and we evaluate AgentSPEX on 7 benchmarks. Finally, we show through a user study that AgentSPEX provides a more interpretable and accessible workflow-authoring paradigm than a popular existing agent framework.
\end{abstract}

\begin{center}
\begin{tabular}{ll}
\faGithub\hspace{0.3em} \textbf{Code} & \url{https://github.com/ScaleML/AgentSPEX} \\
\end{tabular}
\end{center}

\section{Introduction}
AI agents have continued to achieve impressive results on complex tasks ranging from resolving real-world GitHub issues~\citep{jimenez2024swebench, yang2024sweagent} to performing scientific research~\citep{lu2024ai, yu-etal-2025-tinyscientist}. Naturally, a rich ecosystem of open- and closed-source frameworks has emerged in tandem to support agent development, including multi-agent conversation systems~\citep{wu2023autogen}, multi-agent collaboration frameworks~\citep{hong2024metagpt}, software development agent frameworks~\citep{yang2024sweagent, wang2024openhands, anthropic2026claudecode}, among others.

\begin{figure}[t]
\centering
\includegraphics[width=\textwidth]{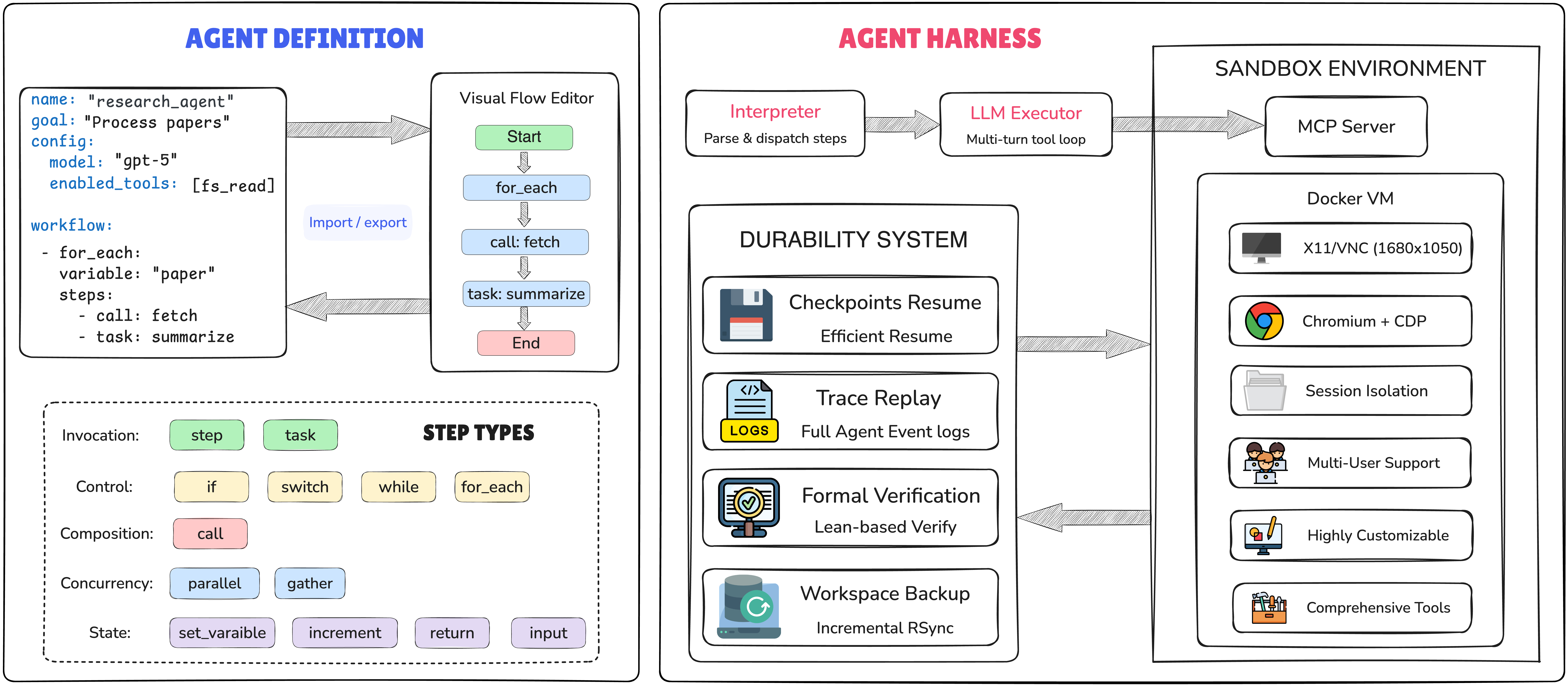}
\caption{An overview of the AgentSPEX Architecture.}
\label{fig:system-architecture}
\vspace{-0.5em}
\end{figure}

Many agent frameworks today adopt a ReAct-style prompting approach~\citep{yao2023react} in which, for a given task, their instructions are specified in a single system prompt–instruction pair. Under such a paradigm, an agent reactively executes a sequence of tool calls and reasoning steps conditioned on its growing conversation history~\citep{Du_2026}. Although this prompting approach is relatively simple to implement in practice, it relies heavily on the inherent capabilities of the chosen foundation model. When compared to more structured problem decomposition and workflow-style approaches, reactive prompting may underperform on longer-horizon tasks, especially those requiring branching and iteration, in terms of performance, cost, reproducibility, and controllability~\citep{yihan2026pseudoactleveragingpseudocodesynthesis, delrosario2025architectingresilientllmagents}. 

To impose more granular control and structure on agent behavior, agent building frameworks such as LangGraph~\citep{langgraph2024}, DSPy~\citep{khattab2024dspy}, and CrewAI~\citep{crewai} provide libraries for defining agent workflows in Python, with features such as stateful graphs with branching and memory, optimizable prompt pipelines, and multi-agent role-based teams. While these frameworks offer greater control over agent execution logic and orchestration, their workflows are highly coupled with Python programming, resulting in steep learning curves and making agents more difficult to modify, maintain, and share with non-programmers.

To address these limitations, we introduce AgentSPEX, an Agent SPecification and EXecution Language, with YAML syntax for specifying agent workflows with explicit control flow and modular structure, along with a customizable agent harness. AgentSPEX supports typed steps, branching and loops, parallel execution, reusable submodules (subagents), and explicit context management, granting users precise control over both agent behavior and context visibility. Figure~\ref{fig:system-architecture} illustrates the full architecture of our agent framework.

Our design philosophy is guided by two principles: AgentSPEX should be expressive enough to capture common agent invocation patterns without requiring modifications to execution source code, and it should remain simple and accessible enough for users to author, inspect, and modify agent behavior with minimal overhead. AgentSPEX features:

\begin{enumerate}[nosep,leftmargin=*]
    \item \textbf{AgentSPEX as the executable specification}, in which agent workflows are expressed in declarative, human-readable YAML files.

    \item \textbf{Unified submodule abstraction}, in which skills and agents are both represented as workflows and can be freely composed, simplifying the development of modular, multi-level agent systems.

    \item \textbf{Explicit conversation history management}, giving users direct control over what context each step receives, improving performance, cost-efficiency, and controllability.
    
    \item \textbf{Agent Harness}, an execution environment that provides tool access, a sandboxed virtual environment, and support for state checkpointing, trajectory logging, and replay and resume capabilities for long-running workflows.
    
    \item \textbf{Bidirectional visual editor integration}, enabling drag-and-drop workflow construction and modification through synchronized graph and workflow views.
\end{enumerate}

AgentSPEX also includes ready-to-use agents that can be deployed for deep research, scientific research proposal generation, and research advising. Furthermore, we include evaluation results on 7 established benchmarks that span the domains of science, writing, scientific paper understanding, and software engineering. 

\begin{figure}[t]
\centering
\begin{minipage}{0.95\linewidth}
\begin{lstlisting}[language=yaml, basicstyle=\ttfamily\scriptsize]
name: "research_assistant"
goal: "Research a topic and write a summary"

config:
  model: "gpt-5.4"
  enabled_tools: ["web_search", "file_write"]
  
parameters:
    topic: "Enhancing LLM reasoning via RLHF"
    file_path: "outputs/report.md"
    
workflow:
  - task:
      instruction: "Generate a list of search queries for {{topic}}"
      save_as: "search_queries"
  - call:
      module: "modules/search_and_summarize.yaml"  # Invoke submodule
      parameters: 
        queries: "{{search_queries}}"
      save_as: "paper_summary"
  - task:
      instruction: "Write a report at {{file_path}} based on these findings: {{paper_summary}}"
\end{lstlisting}
\end{minipage}
\caption{An example of an AgentSPEX workflow for topic research and summarization.}
\label{fig:workflow-example}
\end{figure}

\section{AgentSPEX Design}
\label{sec:approach}
We next present an overview of AgentSPEX, covering both its design and how to write new agents and workflow files. We first introduce the language constructs and how workflows are specified and executed (\S\ref{sec:language}). Then, we describe how workflows manage state and context across steps, and how they compose reusable submodules (\S\ref{sec:programming-model}). Finally, we present the visual editor for convenient workflow construction and iteration (\S\ref{sec:visual}).

\begin{table}[t]
\centering
\begin{tabular}{@{}llp{5.2cm}@{}}
\toprule
\textbf{Construct} & \textbf{Category} & \textbf{Description} \\
\midrule
\texttt{task} & Invocation & Start a new conversation \\
\texttt{step} & Invocation & Continue a persistent conversation \\
\midrule
\texttt{if} / \texttt{switch} & Control flow & Conditional branching \\
\texttt{while} & Control flow & Loop with configurable iteration limit \\
\texttt{for\_each} & Control flow & Iterate over a list \\
\midrule
\texttt{call} & Composition & Invoke another workflow as a submodule \\
\midrule
\texttt{parallel} / \texttt{gather} & Concurrency & Execute operations concurrently \\
\midrule
\texttt{set\_variable} & State & Assign a value to a context variable \\
\texttt{increment} & State & Increment a numeric variable \\
\texttt{input} & State & Prompt the user for input \\
\texttt{return} & State & Return a value to the calling workflow \\
\bottomrule
\end{tabular}
\caption{Workflow constructs that are supported by AgentSPEX}
\label{tab:constructs}
\end{table}

\subsection{Language Constructs and Workflow Representation}
\label{sec:language}
AgentSPEX features a lightweight vocabulary of primitives that allows users to specify agent workflows, covering the execution patterns commonly required in long-horizon tasks without unnecessary complexity. As seen in Figure~\ref{fig:workflow-example}, each workflow follows a common structure consisting of a name, goal, optional config parameters, and a workflow field consisting of a sequence of operations.

\textbf{Config and parameters} define environment-level settings for the agent harness. Common use cases include restricting agents to specific subsets of tools, registering custom sub-agents available for invocation, specifying initial input parameters, and setting an upper bound on the number of tool calls allowed per step.

\textbf{Workflow operations} define the sequence of operations that the agent executes. Table~\ref{tab:constructs} lists the core set of constructs. The two core invocation types are \texttt{task} and \texttt{step}. A \texttt{task} instantiates a new conversation and allows the agent to begin a sequence of tool calls. A \texttt{step}, on the other hand, supports multi-turn interaction, where the model can make tool calls and receive additional instructions/prompts over multiple turns while maintaining conversation history. The remaining constructs handle control flow, composition, concurrency, and state manipulation.

Because workflows are defined in one or a few self-contained YAML files, they are easy to version-control, diff, and share. Furthermore, because instructions are written in natural language within a structured format, domain experts can author and modify workflows without writing Python or navigating orchestration code.

\subsection{State Management and Composition}
\label{sec:programming-model}

\textbf{Context variables and explicit state control.} Each workflow maintains a set of named context variables. Steps reference these variables using Mustache-style templates (e.g., \texttt{\{\{variable\}\}}) and save their outputs to new variables via \texttt{save\_as}. Context variables can then be passed to subsequent instructions. To illustrate, in Figure~\ref{fig:workflow-example}, the first step produces a list of search queries, which is saved into the variable \texttt{search\_queries} and then passed as a parameter into the instruction of a subsequent step.

\textbf{Step vs.\ task.} A \texttt{task} starts a fresh conversation with no prior history, while a \texttt{step} accumulates conversation history across turns. This gives workflow authors direct control over how information flows between instructions: \texttt{task} is appropriate when a specific intermediate result should be passed forward via context variables, and \texttt{step} is suited for cases where the model needs to reason over multiple rounds of tool use and instructions.

\textbf{Composition through workflows.} AgentSPEX uses a single unified abstraction for composition: any workflow can invoke another workflow as a submodule via the \texttt{call} keyword, passing parameters and receiving a return value. Workflows can also be registered as skills/tools, allowing an agent to dynamically select and invoke them during execution. In Figure~\ref{fig:workflow-example}, the agent iterates over a list of papers, calling a search and summarize submodule for each paper.

\subsection{Visual Authoring and Inspection}
\label{sec:visual}
\begin{figure}[h]
    \centering
    \includegraphics[width=\linewidth]{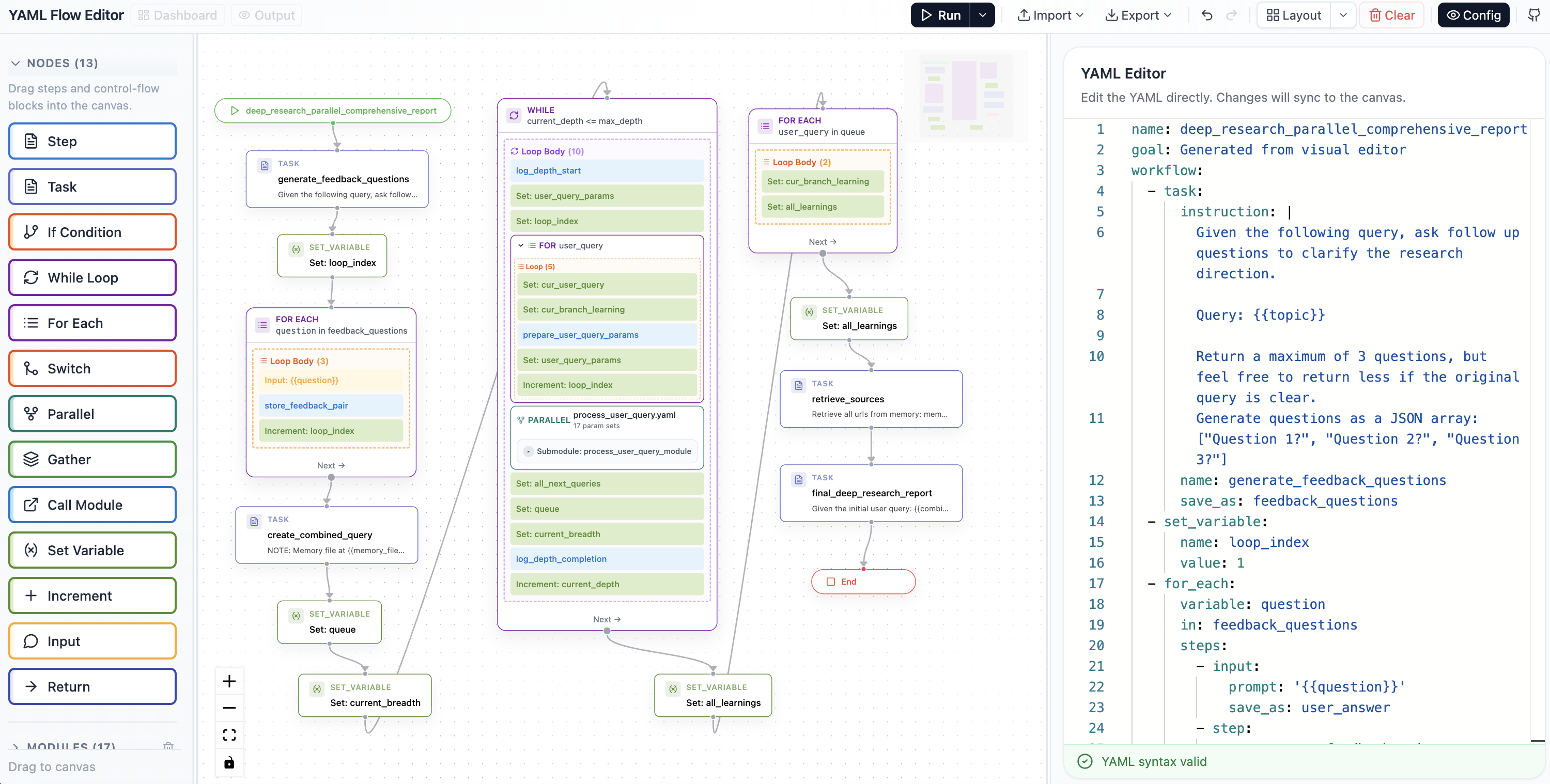}
    \caption{Visual editor interface for a deep research agent implemented with AgentSPEX, showing synchronized graph-based and YAML-based views of a workflow with reusable submodules.}
    \label{fig:yaml-flow-editor-demo}
\end{figure}

We also provide a visual editor for workflow construction and iteration as illustrated in Figure~\ref{fig:yaml-flow-editor-demo}. A workflow is displayed as an interactive flowchart, where each node corresponds to an operation (\texttt{task}, \texttt{step}, \texttt{if}, \texttt{for\_each}, etc.). Users can edit the workflow by adding, removing, or rearranging nodes, or by modifying the workflow directly in a synchronized text panel. Changes in either view are immediately reflected in the other, and the resulting workflow can be executed within the visual editor itself.

\section{Agent Harness}
\label{sec:agent-harness}
We next describe the agent harness that executes AgentSPEX agents and workflows. We begin with the execution engine, including its sandboxed execution environment (\S\ref{sec:exec-engine}). We then describe the observability dashboard (\S\ref{observability}), which supports debugging and real-time monitoring, before turning to the durability mechanisms that support long-running workflows through checkpointing, tracing, replay, and resume (\S\ref{sec:durability}).

\subsection{Execution Engine}
\label{sec:exec-engine}

\textbf{Interpreter}. The interpreter is the entry point for workflow execution. Given a workflow file, it validates the workflow structure, resolves configuration parameters, and expands template variables. The interpreter then iterates through the sequence of operations in the given workflow, dispatching each operation to the appropriate step handler based on its type. For nested constructs (loops, conditionals, submodule calls), it manages recursion and variable scoping. Each operation is assigned a hierarchical step identifier (e.g., 3.2.1 for the first substep of the second iteration of operation 3), which is used for checkpointing and logging.

\textbf{Executor}. The executor implements the interaction loop between the language model and external tools. For each \texttt{step} or \texttt{task} operation, the executor runs a multi-turn loop that terminates when the model returns a response with no tool calls or the configured tool-call or token limit is reached. In each iteration of the multi-turn loop, the executor sends the current message history to the model, executes any tool calls in the response via a Model Context Protocol (MCP;~\citealp{anthropic2024mcp}) client, appends the results to the message history, and continues.

\textbf{Execution Environment}. Each workflow executes within a Docker-based sandbox that provides an isolated environment equipped with browser and file system access, and access to over 50 tools spanning the categories of file operations, web search, code execution, browser automation, among others.

\subsection{Observability}
\label{observability}
To support debugging and real-time monitoring, the agent harness includes a built-in observability dashboard. This dashboard shows live logs of agent actions and intermediate reasoning steps, allowing users to inspect agent behavior at each stage of a workflow. Figure~\ref{fig:dashboard} in Appendix~\ref{app:observability} shows an example of the dashboard during the execution of a SWE-Bench Verified instance.

\subsection{Durability}
\label{sec:durability}
Long-running agent workflows are susceptible to interruptions from sources such as API timeouts, rate limits, and network failures. The durability system provides more robust execution through checkpointing and execution tracing.

\textbf{Checkpointing.} Checkpoints are saved by the harness after each step completes, recording completed step identifiers, current context (all template variable values and prior step outputs), step-level metrics, and the current sandbox state. Execution can then be resumed from any checkpoint. The system restores the saved context, skips completed steps, and reattaches to the original sandbox session, allowing the workflow to continue from where it left off.

\textbf{Execution Tracing and Replay.} The agent harness records a full execution trace for each workflow run, capturing model responses, tool-calling results, and the conversation state at each step. The harness also supports selective trace replay, which addresses a common challenge in workflow development: modifying a single step's instruction typically requires re-executing the entire workflow, including unchanged upstream steps. With selective trace replay, the executor can load a specified number of steps from a prior trace and resume live execution, allowing developers to isolate the effect of a prompt or control-flow change on downstream behavior while holding upstream context constant.

\textbf{Formal Verification.} AgentSPEX makes control flow, variable dependencies, and step boundaries explicit in workflow specifications. This enables the formal verification of both structural and semantic correctness for agent plans and their execution trajectories, based on a predicate system defined in formal languages such as Lean~\citep{moura2021lean} or Isabelle~\citep{paulson1994isabelle}. Such verification can be performed by defining the pre- and post-conditions for each execution step. We further discuss the potential for formal verification in Appendix~\ref{discussion:formal}.

\section{Demos and Evaluation}
\label{sec:evaluation}
To demonstrate the effectiveness of AgentSPEX, we provide three ready-to-use agents for deep research, scientific research proposal generation, and research advising. Additionally, we evaluate workflows written in AgentSPEX on 7 established benchmarks that span the domains of science, mathematics, writing, scientific paper understanding, and software engineering. We provide results comparing the performance of established baselines and existing agent frameworks with workflows that utilize AgentSPEX for each respective benchmark. Videos demonstrating each agent are also provided, along with the workflow files used\footnote{\url{https://anonymous.4open.science/r/AgentSPEX_video_demo-DC49/}}.

\subsection{Agent Demos}
\label{sec:demos}
\textbf{Deep Research} takes a user query as input and generates a comprehensive report in Markdown, mirroring popular closed-source deep research offerings such as OpenAI Deep Research~\citep{openai_deep_research_2025} and Gemini Deep Research~\citep{google_gemini_deep_research}. As illustrated in Figure~\ref{fig:yaml-flow-editor-demo}, the workflow implements a multi-level searching strategy governed by two configurable parameters: breadth, which controls the number of queries explored at each level, and depth, which controls how many levels of follow-up queries are run. At each level, the agent generates targeted search queries, executes search queries in parallel, and extracts intermediate findings.

\textbf{AI Scientist} takes an initial user intent and generates a novel academic research proposal through a two-stage pipeline. The workflow used is based on the implementation provided in~\cite{yu-etal-2025-tinyscientist}. In Stage~1 (Thinker), the system performs safety classification, generates search queries, retrieves related work via OpenAlex~\citep{priem2022openalex}, and iteratively generates and refines research ideas until a sufficiently novel proposal emerges. In Stage~2 (Writer), the system composes paper sections sequentially while dispatching parallel citation workers to locate and insert references.

\textbf{AI Advisor} is a research advising agent that takes a research proposal or paper as input and produces a rubric-based review along with actionable feedback and suggestions. The workflow structure follows the implementation described in~\cite{liu2025guidescalableadvisingresearch}. The agent first parses the input document and segments it into semantic chunks. Then, it summarizes each chunk into structured JSON before synthesizing the results into a comprehensive review covering summary, novelty, significance, soundness, strengths, weaknesses, and recommendations.

\begin{table*}[t]
\centering
\small
\setlength{\tabcolsep}{6pt}
\renewcommand{\arraystretch}{1.10}
\begin{adjustbox}{width=\textwidth}
\begin{tabular}{@{} l l l c @{}}
\toprule
\textbf{Agent} & \textbf{Model} & \textbf{Domain} & \textbf{Score} \\

\midrule
\rowcolor{blue!8}
\multicolumn{4}{c}{\textbf{SciBench}~\citep{wang2024scibenchevaluatingcollegelevelscientific}} \\
CoT & GPT-5 & Science & 85.92\% \\
ReAct & GPT-5 & Science & 87.79\% \\
AgentSPEX (Ours) & GPT-5 & Science & \textbf{90.61\%} \\

\midrule
\rowcolor{blue!8}
\multicolumn{4}{c}{\textbf{StemEZ}~\citep{wang2024mmlu}} \\
CoT & GPT-5 & Science & 82.87\% \\
ReAct & GPT-5 & Science & 84.72\% \\
AgentSPEX (Ours) & GPT-5 & Science & \textbf{86.57\%} \\

\midrule
\rowcolor{blue!8}
\multicolumn{4}{c}{\textbf{ChemBench}~\citep{mirza2025framework}} \\
CoT & GPT-5$^*$ & Science & 78.90\% \\
ReAct & GPT-5$^*$ & Science & 77.80\% \\
AgentSPEX (Ours) & GPT-5$^*$ & Science & \textbf{83.30\%} \\

\midrule
\rowcolor{blue!8}
\multicolumn{4}{c}{\textbf{AIME 2025}~\citep{maa_aime_2026}} \\
CoT~\citep{openai2025gpt5} & GPT-5 (without tools) & Mathematics & 94.60\% \\
CoT~\citep{openai2025gpt5} & GPT-5 (with Python) & Mathematics & 99.60\% \\
AgentSPEX (Ours) & GPT-5 & Mathematics & \textbf{100.0\%} \\

\midrule
\rowcolor{blue!8}
\multicolumn{4}{c}{\textbf{ELAIPBench}~\citep{dai2026elaipbenchbenchmarkexpertlevelartificial}} \\
CoT & GPT-5$^*$ & Paper Understanding & 37.22\% \\
ReAct & GPT-5$^*$ & Paper Understanding & 33.80\% \\
AgentSPEX (Ours) & GPT-5$^*$ & Paper Understanding & \textbf{43.70\%} \\

\midrule
\rowcolor{blue!8}
\multicolumn{4}{c}{\textbf{WritingBench}~\citep{wu2025writingbench}} \\
CoT & Claude-Sonnet-4.5-Thinking & Writing & 79.90\% \\
ReAct & Claude-Sonnet-4.5-Thinking & Writing & 80.30\% \\
AgentSPEX (Ours) & Claude-Sonnet-4.5-Thinking & Writing & \textbf{81.00\%} \\

\midrule
\rowcolor{blue!8}
\multicolumn{4}{c}{\textbf{SWE-Bench Verified}~\citep{jimenez2024swebench}} \\
mini-SWE-agent~\citep{yang2024sweagent} & Claude-Opus-4.5$^*$/4.6$^*$ & Software Engineering & 76.20\% \\
Live-SWE-agent~\citep{xia2025livesweagentsoftwareengineeringagents} & Claude-Opus-4.5$^*$/4.6$^*$ & Software Engineering & 74.60\% \\
AgentSPEX (Ours) & Claude-Opus-4.5$^*$/4.6$^*$ & Software Engineering & \textbf{77.10\%} \\

\bottomrule
\end{tabular}
\end{adjustbox}
\caption{Evaluation results on seven different benchmarks. SWE-Bench Verified results are reported as the average performance of Claude-Opus-4.5 and Claude-Opus-4.6. $^*$Denotes use of high-reasoning effort.}
\label{tab:benchmark-results}
\end{table*}

\subsection{Benchmarks and Results}
\label{sec:benchmarks} 
We evaluated AgentSPEX on 7 benchmarks spanning a wide range of domains as summarized in Table~\ref{tab:benchmark-results}. For most benchmarks, we compare against two baselines: a chain-of-thought (CoT;~\citealp{wei2022chain}) baseline that provides the model with all necessary information in a single prompt, and a ReAct~\citep{yao2023react} baseline that includes the AgentSPEX workflow in the initial prompt but allows the agent to interpret it reactively and call tools it deems necessary without enforced step-by-step execution. AgentSPEX workflows are written manually with the help of coding assistants. We primarily used GPT-5~\citep{openai2025gpt5} for our experiments apart from Claude-Sonnet-4.5-Thinking~\citep{anthropic2025claudesonnet45} for WritingBench and Claude-Opus-4.5/4.6~\citep{anthropic2026claudeopus46} for SWE-Bench Verified. 

\begin{itemize}
    \item For \textbf{science}, we use three benchmarks that test multi-step quantitative reasoning on STEM problems. SciBench~\citep{wang2024scibenchevaluatingcollegelevelscientific} contains college-level scientific problems; we use its chemistry subsets to test multi-step quantitative and symbolic reasoning. StemEZ~\citep{wang2024mmlu} is a physical chemistry subset of MMLU-Pro requiring complex formula derivations and multi-step calculations. ChemBench~\citep{mirza2025framework} covers chemical knowledge and reasoning with over 2,700 question--answer pairs spanning undergraduate and graduate curricula.
    \item For \textbf{mathematics}, AIME 2025~\citep{maa_aime_2026} is based on the American Invitational Mathematics Examination, a competition-level exam with 30 problems whose answers are integers between 0 and 999.
    \item For \textbf{paper understanding}, ELAIPBench~\citep{dai2026elaipbenchbenchmarkexpertlevelartificial} contains 403 multiple-choice questions derived from 137 AI research papers and tests deep comprehension over full-length academic documents rather than shallow fact retrieval.
    \item For \textbf{generative writing}, WritingBench~\citep{wu2025writingbench} spans 6 major domains and 100 subdomains, evaluating whether models can satisfy requirements involving style, format, and length across creative, persuasive, informative, and technical writing tasks.
    \item For \textbf{software engineering}, SWE-Bench Verified~\citep{jimenez2024swebench} contains 500 human-verified GitHub issues where an agent must produce a patch that resolves a bug or feature request while preserving existing functionality; correctness is determined by hidden unit tests.
\end{itemize}

Full evaluation details, including subset selection and evaluation methods, are provided in Appendix~\ref{appendix:evaluation-details}.

\paragraph{Overall results.} As shown in Table~\ref{tab:benchmark-results}, AgentSPEX achieves the highest score among the compared approaches on all 7 benchmarks. On the science benchmarks, it improves over the stronger baseline by 2.8\% on SciBench, 1.9\% on StemEZ, and 5.5\% on ChemBench. On ELAIPBench, the improvement over CoT is 6.5\%. On AIME 2025, AgentSPEX achieves a perfect score of 100\%, and on SWE-Bench Verified it scores 77.1\%, outperforming both mini-SWE-agent (76.2\%) and Live-SWE-agent (74.6\%). WritingBench shows a more modest gain (81.0\% vs.\ 80.3\%).
 
\paragraph{Discussion.} An interesting pattern emerges from comparing the ReAct and AgentSPEX results. Both approaches use the same workflow, but the ReAct baseline allows the model to interpret the workflow freely while AgentSPEX enforces step-by-step execution. On ELAIPBench, the ReAct baseline scores 3.4\% below CoT (33.8\% vs.\ 37.2\%), and on ChemBench it underperforms CoT by 1.1\%. This can be attributed to the fact that including a structured workflow in the prompt without enforcement adds complexity to the model’s task, as the model must simultaneously interpret the workflow structure and reason about the problem. Enforcing execution through the harness, on the other hand, may alleviate this burden by offloading control flow logic to the interpreter.
 
The improvements tend to be larger on benchmarks that involve processing substantial input material or coordinating multiple reasoning steps. ChemBench (+5.5\%) and ELAIPBench (+6.5\%) both require the agent to reason over extended problem statements or full-length research papers, and may benefit from AgentSPEX's explicit context management, which controls what information each step receives. Without such mechanisms, the ReAct paradigm must carry the full conversation history forward across all reasoning steps, which may lead to context degradation on longer-horizon tasks.

\subsection{User Study}
\textbf{Setup.} We conducted a user study using Google Forms with 23 participants. Each participant was shown two workflows that implement the same agentic behavior. One is written in AgentSPEX and the other is written in LangGraph. Participants were then asked to rate their level of experience in agent development and their preferences between the two workflow styles along the axes of interpretability and preference for adoption. Finally, participants were required to list advantages and disadvantages of each workflow. Additional details regarding the setup of the user study are provided in Appendix~\ref{app:user-study}.

\textbf{Qualitative Results.}
Participants generally favored AgentSPEX for its readability, clarity of prompting, and ease of use when creating a new agent workflow from scratch. In contrast, most participants preferred LangGraph for constructing \textit{complex, multi-step} agent workflows. Open-ended responses reinforced these patterns: AgentSPEX was described as \textit{``accessible to non-coders''} and \textit{``easier to understand,''} whereas LangGraph was characterized as \textit{``customizable''} and \textit{``more rigorous.''} Together, these findings suggest that AgentSPEX is perceived as more approachable, but that participants were less confident in its ability to support more complex workflows. Our demos in Section~\ref{sec:demos} help address these concerns by demonstrating AgentSPEX's ability to support complex, production-ready agents.

\section{Related Work}
\label{sec:related}

\begin{table}[t]
\centering
\small
\resizebox{\linewidth}{!}{%
\begin{tabular}{lccc}
\toprule
Approach & Natural Language & Explicit Context & Visual Editor \\
\midrule
AutoGen~\citep{wu2023autogen}  & \ding{55} & \ding{55} & \ding{55} \\
DSPy~\citep{khattab2024dspy}   &  \ding{55} & \ding{55} & \ding{55} \\
CrewAI~\citep{crewai}          & Partial & \ding{55} & \ding{55} \\
LangGraph w/ LangFlow~\citep{langflow2026} & \ding{55} & \ding{55} & \ding{51} \\
n8n~\citep{n8n2026} & \ding{55} & \ding{55} & \ding{51} \\
ADL~\citep{zeng2025adldeclarativelanguageagentbased} & \ding{51} & \ding{55} & \ding{55} \\
PDL~\citep{vaziri2024pdldeclarativepromptprogramming} & \ding{51} & Partial & \ding{55} \\
\textbf{AgentSPEX (Ours)} &  \ding{51} & \ding{51} & \ding{51} \\
\bottomrule
\end{tabular}%
}
\caption{Comparison of agent building frameworks. \textbf{Natural Language}: supports specifying workflow logic directly in natural language. \textbf{Explicit Context}: supports explicit, user-controlled context injection through variables or structured inputs. \textbf{Visual Editor}: provides a graphical interface for composing or editing workflows.}
\label{tab:comparison}
\end{table}

As AI agents have grown in both capability and popularity, so too has the range of frameworks available for building them. Table~\ref{tab:comparison} summarizes key differences among several open-source agent-building frameworks, including their support for visual no-code editors, implementation style, and context management. Prior work has also highlighted the importance of robust agent harnesses with sandboxed environments that give agents access to general computer-use tools~\citep{cheng2026llminsandboxelicitsgeneralagentic}.

This trend extends beyond open-source frameworks. With the emergence of closed-source agent platforms such as Codex CLI skills~\citep{openai2026codexskills} and Claude Code skills~\citep{anthropic2026claudecode}, users can now define reusable skills or submodules directly in natural language. Although these approaches are often fast to implement in practice, they shift control flow, state management, and execution semantics to the foundation model at runtime, which can limit reproducibility and reduce user control.

These trade-offs become especially important when considering one of the central challenges facing modern AI agents: context rot~\citep{hong2025contextrot} and the broader performance degradation that occurs as contexts grow longer~\citep{liu2024lost, du-etal-2025-context}. As a result, a core design decision for any agent framework is how to manage the context available to the agent at each step. Memory management frameworks such as MemGPT~\citep{packer2023memgpt} address this through hierarchical memory management, while ACON~\citep{kang2025acon} applies context compression to reduce token consumption in long-horizon agents.

\section{Conclusion and Future Work}
\label{sec:conclusion}
AgentSPEX provides a structured and controllable framework for specifying LLM-agent workflows. By making control flow, composition, and context management explicit, it enables agents that are easier to author, inspect, and maintain. We hope this work contributes to the development of more reliable, modular, and accessible agent systems. 

Promising directions for future work include formal verification of agent execution, training models to automatically write and use workflows, incorporating end-to-end agentic training pipelines into the framework, and additional support for multi-agent orchestration. Furthermore, advancing the framework's support for long-context reasoning and longer-horizon tasks through more robust context-compression methods and more expressive abstractions for multi-agent orchestration remains an exciting direction.

\bibliography{main}
\bibliographystyle{colm2026_conference}

\appendix

\section{Evaluation Details}
\label{appendix:evaluation-details}
 
We evaluate on seven diverse benchmarks spanning five domains, summarized in Table~\ref{tab:eval_summary}. All evaluations use pass@1 accuracy and evaluate on the full test set unless otherwise specified. Please refer to our released code for the exact workflows and agent harness configurations.
 
\begin{table}[h]
\centering
\small
\setlength{\tabcolsep}{4pt}
\begin{tabular}{llcl}
\toprule
\textbf{Domain} & \textbf{Benchmark} & \textbf{\# Problems} & \textbf{Evaluation} \\
\midrule
Software Engineering & SWE-Bench Verified \citep{jimenez2024swebench} & 500 & sb-cli \\
Mathematics & AIME 2025 \citep{maa_aime_2026} & 30 & Math-Verify \\
\multirow{3}{*}{Chemistry} & ChemBench \citep{mirza2025framework} & 90 & Exact Match \\
                            & SciBench \citep{wang2024scibenchevaluatingcollegelevelscientific} & 213 & Exact Match \\
                            & MMLU-Pro Stemez \citep{wang2024mmlu} & 216 & Exact Match \\
Generative Writing & WritingBench \citep{wu2025writingbench} & 120 & LLM-as-Judge \\
Paper Understanding & ELAIPBench \citep{dai2026elaipbenchbenchmarkexpertlevelartificial} & 403 & Exact Match \\
\bottomrule
\end{tabular}
\caption{Summary of evaluation benchmarks.}
\label{tab:eval_summary}
\end{table}
 
\subsection{SWE-Bench Verified}
\label{app:swe-bench-analysis}
\citet{chen2023chatgpt} documented significant variation in LLM outputs over time, motivating the need for rapid agent iteration. Therefore, an important but often overlooked dimension of agent framework evaluation is model-version robustness: whether an agent system maintains or improves performance when the underlying foundation model is upgraded. We observe that tightly coupled agent systems where prompts, control flow, and orchestration logic are interleaved within Python source code can be sensitive to changes in model behavior, as even minor shifts may require coordinated modifications across both prompt content and execution code.
 
\paragraph{Setup.}
We evaluate all three agent systems on SWE-Bench Verified using both Claude-Opus-4.5 and Claude-Opus-4.6 with reasoning effort set to high. Table \ref{tab:swe_model_robustness} reports the per-model results; the averages are reported in Table \ref{tab:benchmark-results}.

\begin{table}[h]
\centering
\begin{tabular}{lccc}
\toprule
\textbf{Agent} & \textbf{Claude-Opus-4.5} & \textbf{Claude-Opus-4.6} & \textbf{$\Delta$} \\
\midrule
mini-SWE-agent & 76.8\% & 75.6\% & $-$1.2 \\
Live-SWE-agent & 78.0\% & 71.2\% & $-$6.8 \\
AgentSPEX (Ours) & 77.2\% & 77.0\% & $-$0.2 \\
\bottomrule
\end{tabular}
\caption{SWE-Bench Verified results broken down by model version. All runs use high-reasoning effort with temperature set to 1.0.}
\label{tab:swe_model_robustness}
\end{table}

\paragraph{Reproducibility note.}
For Live-SWE-agent~\citep{xia2025livesweagentsoftwareengineeringagents} using Claude-Opus-4.5, we averaged the results reported in the paper with a locally reproduced score of 78.0\%. 
 
\subsection{Science}
\begin{itemize}[leftmargin=*,itemsep=2pt]
    \item \textbf{ChemBench}~\citep{mirza2025framework} is an automated framework for evaluating chemical knowledge and reasoning, containing 2,788 question--answer pairs across 9 domains. We randomly sample 10 questions from each domain, yielding a 90-question subset. Evaluation uses exact match.
    \item \textbf{SciBench}~\citep{wang2024scibenchevaluatingcollegelevelscientific} is a collegiate-level scientific problem-solving benchmark featuring open-ended, free-response questions from widely-used textbooks. We use the chemistry subsets: Atkins' Physical Chemistry ($N{=}101$), Chemistry by McMurry \& Fay ($N{=}33$), Properties of Matter ($N{=}47$), and Quantum Chemistry ($N{=}32$), totaling 213 problems. Evaluation uses exact match against reference solutions.
    \item \textbf{MMLU-Pro Stemez}~\citep{wang2024mmlu} provides STEM problems sourced from the Stemez website (\url{https://stemez.com}). We use the Physical Chemistry subset containing 216 problems. Evaluation uses exact match.
\end{itemize}
 
\subsection{Generative Writing}
WritingBench~\citep{wu2025writingbench}  is a comprehensive benchmark for generative writing that includes queries spanning 6 major domains and 100 subdomains. We randomly sample 20 questions from each of the 6 domains, yielding a 120-question subset. Evaluation follows the official protocol using LLM-as-Judge, where each response is assessed by 5 instance-specific criteria on a 10-point scale via the provided critic model.

\section{Observability}
\label{app:observability}
\begin{figure}[h]
    \centering
    \includegraphics[width=\linewidth]{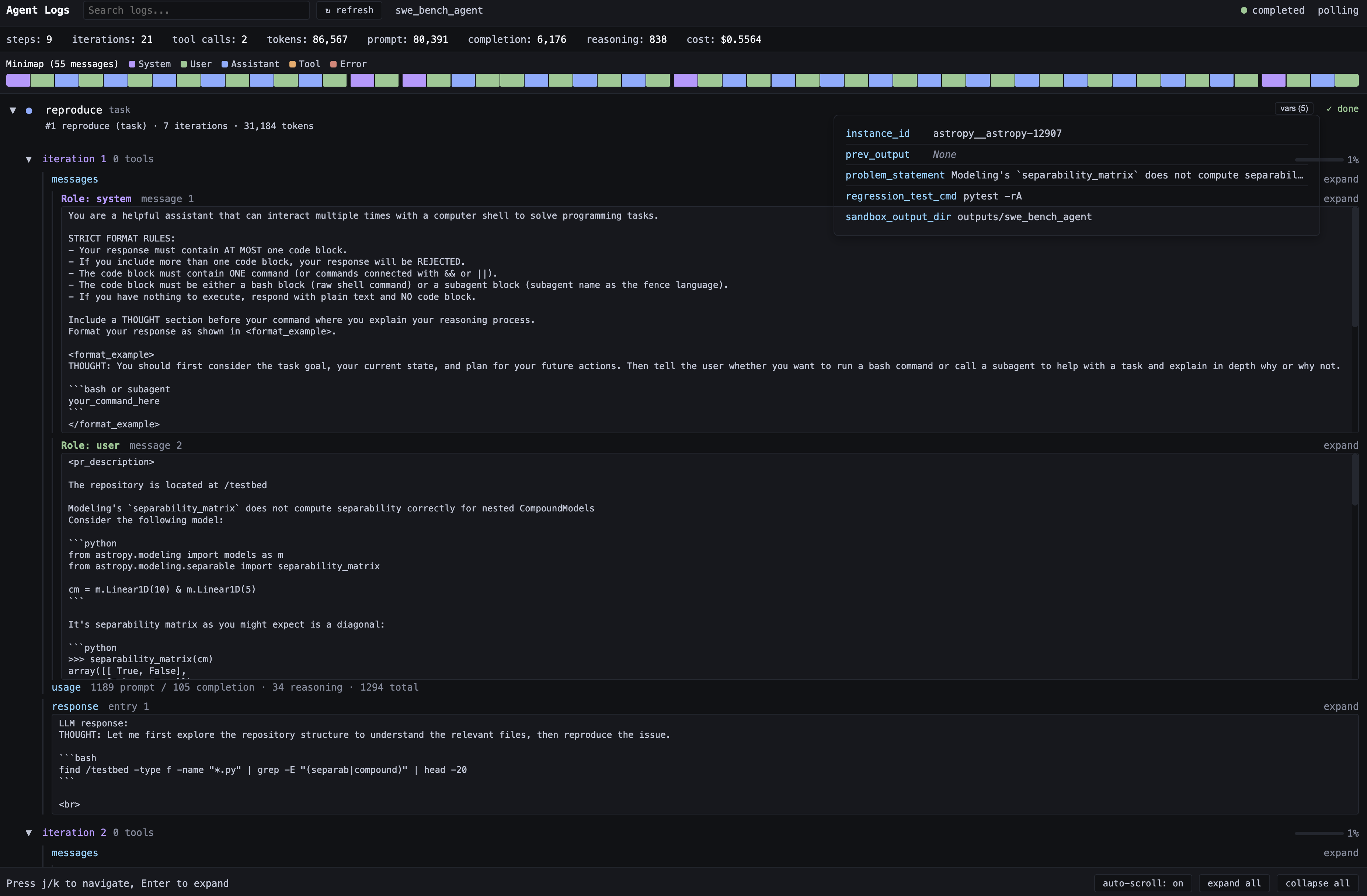}
    \caption{Example of the log viewer for a SWE-Bench Verified instance.}
    \label{fig:dashboard}
\end{figure}

We provide a screenshot of the observability dashboard in Figure~\ref{fig:dashboard} for reference.

\section{Formal verification of AgentSPEX}\label{discussion:formal}
\label{app:verification}
A notable advantage of AgentSPEX's declarative structure is that it enables formal verification of agent workflows, a capability that is difficult to achieve with imperative frameworks where control flow is embedded in general-purpose code. Because AgentSPEX makes control flow, variable dependencies, and step boundaries explicit in the YAML specification, both the agent plan and its execution trajectory become amenable to static and runtime verification. We demonstrate the potential for an AgentSPEX plan to be formally verified in the sense of trajectory. We demonstrate this by verifying the \texttt{extract\_single\_citation\_module} in our AI scientist framework. The original task plan is demonstrated in Figure~\ref{fig:verifiable_plan}. Based on the LLM instruction in the agent plan, we can infer the property each variable must satisfy before execution for verification, as demonstrated in Figure~\ref{fig:properity_of_var}.
Based on the task plan, we performed a reasoning and obtained the trajectory demonstrated in Figure~\ref{fig:verifiable_plan_trajectory}. The trajectory can be verified through formal verification or external tool verification. The example for verification is demonstrated in Figure~\ref{fig:traj_verification}. 
Using the verification, we can ensure whether the agent plan is properly executed and presents to be a promising future direction to regulate the behavior of agentic systems.

\begin{figure}[t]
\centering
\vspace{-0.4in}
\begin{lstlisting}[language=yaml, basicstyle=\ttfamily\scriptsize]
name: "extract_single_citation_module"
goal: "Extract citation information from a single downloaded paper file"

# Parameters from parent workflow
parameters:
    url: "${URL}"
    file_path: "${FILE_PATH}"

workflow:
    # ========================================
    # Step 1: Extract citation metadata
    # ========================================
    - step:
        name: "extract_paper_title"
        instruction: |
            NOTE: Use the EXACT path provided below, DO NOT add any parent directories !!!

            Read the first 3000 bytes from the paper file at: {{file_path}}

            Extract ONLY the title of the paper. The title is usually:
            - At the top of the first page
            - In a larger or bold font
            - Above the author names

            Return ONLY the extracted title as a single line of text with no extra commentary or formatting.
        save_as: "paper_title"

    - step:
        name: "extract_bibtex_by_tool"
        instruction: |
            Use get_bibtex_from_url tool to retrieve the bibtex using the given url: {{url}} and the paper title: {{paper_title}}

            ONLY return the content of the bibtex with no extra commentary.
        save_as: "bibtex_citation"

    - step:
        name: "extract_abstract_by_tool"
        instruction: |
            Use get_abstract_from_url tool to retrieve the bibtex using the given url: {{url}} and the paper title: {{paper_title}}

            ONLY return the content of the abstract with no extra commentary, if no abstract retreived, ONLY return None.
        save_as: "abstract"
    
    - step:
        name: "prepare_return_dict"
        instruction: |
            Return as python compatible JSON dict:

            {
                "bibtex": "{{bibtex_citation}}",
                "file_path": "{{file_path}}",
                "title": "{{paper_title}}",
                "abstract": "{{abstract}}"
            }

            ONLY return the python dict with no extra commentary.
        save_as: "return_dict"

    - return: "return_dict"
\end{lstlisting}
\caption{Example \texttt{extract\_single\_citation\_module} YAML plan for formal verification}
\label{fig:verifiable_plan}
\end{figure}

\begin{figure}[t]
\centering
\vspace{-0.4in}
\begin{lstlisting}[basicstyle=\ttfamily\scriptsize]
STATIC SEMANTIC VERIFICATION: extract_single_citation (7 nodes, 6 edges)
 
Node 1 (extract_paper_title)
  Preconditions:   fs_read: toolExists
                   file_path: isValidFilePath
  Postconditions:  paper_title: isNonEmptyString                                          OK
 
Node 2 (extract_bibtex_by_tool)
  Preconditions:   paper_title: isNonEmptyString                              [from Node 0]
                   url: isValidURL
  Postconditions:  bibtex_citation: isValidBibtex                                         OK
 
Node 3 (extract_abstract_by_tool)
  Preconditions:   paper_title: isNonEmptyString                              [from Node 0]
                   url: isValidURL
  Postconditions:  abstract: nameExists                                                   OK
 
Node 4 (prepare_return_dict)
  Preconditions:   abstract: isNonEmptyString
                   bibtex_citation: isValidBibtex
                   file_path: isValidFilePath
                   paper_title: isNonEmptyString
  Postconditions:  return_dict: isValidJson
                   return_dict: matchesJsonSchema({bibtex, file_path, title, abstract})   OK
 
Node 5 (return_result)
  Preconditions:   return_dict: isValidJson
                   return_dict: matchesJsonSchema({bibtex, file_path, title, abstract})   OK
 
Graph verification: PASSED (7/7 nodes)
\end{lstlisting}
\caption{Properties of variables inferred from the YAML task plan}
\label{fig:properity_of_var}
\end{figure}

\begin{figure}[t]
\centering
\begin{lstlisting}[basicstyle=\ttfamily\scriptsize]
model: claude-4.6-opus
 
Step 1: extract_paper_title                                                        [completed]
  context:  {url, file_path}
  tool:     extract_text_from_file(file_path="/workspace/...11769v2.pdf") -> ok
  output:   paper_title = "GAR: GENERATIVE ADVERSARIAL REINFORCEMENT LEARNING ..."
  tokens:   16,864
 
Step 2: extract_bibtex_by_tool                                                     [completed]
  context:  {url, file_path, paper_title}
  tool:     get_bibtex_from_url(url="https://arxiv.org/abs/2510.11769", title=...) -> ok
  output:   bibtex_citation = "@misc{https://doi.org/10.48550/arxiv.2510.11769, ...}"
  tokens:   16,463
 
Step 3: extract_abstract_by_tool                                                   [completed]
  context:  {url, file_path, paper_title, bibtex_citation}
  tool:     get_abstract_from_url(url="https://arxiv.org/abs/2510.11769", title=...) -> ok
  output:   abstract = "Solving math problems through verifiable languages ..."
  tokens:   16,685
 
Step 4: prepare_return_dict                                                        [completed]
  context:  {abstract, bibtex_citation, file_path, paper_title}
  output:   return_dict = {"bibtex": "...", "file_path": "...", "title": "...", "abstract": "..."}
  tokens:   8,981
 
Step 5: return(return_dict)                                                        [completed]
 
Workflow total: 58,993 tokens | Cost: $0.025
\end{lstlisting}
\caption{Example \texttt{extract\_single\_citation\_module} YAML plan for formal verification}
\label{fig:verifiable_plan_trajectory}
\end{figure}

\begin{figure}[t]
\centering
\vspace{-0.4in}
\begin{lstlisting}[basicstyle=\ttfamily\scriptsize]
DYNAMIC EXECUTION VERIFICATION: extract_single_citation
 
Node 1 (extract_paper_title)                                                      [completed]
  Preconditions:
    file_path : isValidFilePath                                           -> PASS [tool]
  Postconditions:
    paper_title : isNonEmptyString                                        -> PASS [formal]
 
Node 2 (extract_bibtex_by_tool)                                                   [completed]
  Preconditions:
    paper_title : isNonEmptyString                                        -> PASS [formal]
    url : isValidURL                                                      -> PASS [tool]
  Postconditions:
    bibtex_citation : isValidBibtex                                       -> PASS [tool]
 
Node 3 (extract_abstract_by_tool)                                                 [completed]
  Preconditions:
    paper_title : isNonEmptyString                                        -> PASS [formal]
    url : isValidURL                                                      -> PASS [tool]
  Postconditions:
    abstract : nameExists                                                 -> PASS [formal]
 
Node 4 (check_abstract == None)                                                   [completed]
  Preconditions:
    abstract : nameExists                                                 -> PASS [formal]
  Postconditions:
    abstract : isNonEmptyString                                           -> PASS [formal]
 
Node 5 (prepare_return_dict)                                                      [completed]
  Preconditions:
    abstract : isNonEmptyString                                           -> PASS [formal]
    bibtex_citation : isValidBibtex                                       -> PASS [tool]
    file_path : isValidFilePath                                           -> PASS [tool]
    paper_title : isNonEmptyString                                        -> PASS [formal]
  Postconditions:
    return_dict : isValidJson                                             -> PASS [tool]
    return_dict : matchesJsonSchema({bibtex, file_path, title, abstract}) -> PASS [tool]
 
Node 6 (return_result)                                                            [completed]
  Preconditions:
    return_dict : isValidJson                                             -> PASS [tool]
    return_dict : matchesJsonSchema({bibtex, file_path, title, abstract}) -> PASS [tool]
 
ALL NODES VERIFIED
\end{lstlisting}
\caption{Example of formal verification of trajectory}
\label{fig:traj_verification}
\end{figure}

\section{User Study}
\label{app:user-study}
We provide the questions used in the user study in Table~\ref{tab:survey-questions}. Survey participants generally all had prior programming experience, but had varied levels of experience building agents.

\begin{table}[h]
  \centering
  \small
  \begin{tabular}{@{}llp{8.2cm}@{}}
    \toprule
    \textbf{ID} & \textbf{Type} & \textbf{Question} \\
    \midrule

    Background & Background &
    How much agent development experience do you have? \\
    \addlinespace

    Q0 & Comprehension &
    What task are the agent declarations doing? \\
    \addlinespace

    Q1 & Preference &
    Which implementation is easier to read and understand? \\
    \addlinespace

    Q2 & Preference &
    Which approach would you find easier to start building a new agent
    workflow from scratch? \\
    \addlinespace

    Q3 & Preference &
    Which approach makes it easier to find and understand what LLM
    prompts are being used? \\
    \addlinespace

    Q4 & Preference &
    If you were building a complex multi-step agent workflow, which
    approach would you choose? \\
    \addlinespace

    Q5 & Open-ended &
    What do you see as the main advantages of AgentSPEX? \\
    \addlinespace

    Q6 & Open-ended &
    What do you see as the main weaknesses of AgentSPEX? \\
    \addlinespace

    Q7 & Open-ended &
    What do you see as the main advantages of LangGraph? \\
    \addlinespace

    Q8 & Open-ended &
    What do you see as the main weaknesses of LangGraph? \\
    \bottomrule
  \end{tabular}
  \caption{User study survey questions. The labels for the two agent frameworks are anonymized during the survey.}
  \label{tab:survey-questions}
\end{table}

\section{LLM Use Statement}
LLMs and coding agents were primarily used as an aid during the coding of our agent framework as well as in the writing of our workflow files. Furthermore, LLMs were used to find related works and polish the writing of the paper.
\end{document}